\documentclass[letterpaper]{article} % DO NOT CHANGE THIS
\usepackage{aaai25}  % DO NOT CHANGE THIS
\usepackage{times}  % DO NOT CHANGE THIS
\usepackage{helvet}  % DO NOT CHANGE THIS
\usepackage{courier}  % DO NOT CHANGE THIS
\usepackage[hyphens]{url}  % DO NOT CHANGE THIS
\usepackage{graphicx} % DO NOT CHANGE THIS
\usepackage{natbib}  % DO NOT CHANGE THIS AND DO NOT ADD ANY OPTIONS TO IT
\usepackage{caption} % DO NOT CHANGE THIS AND DO NOT ADD ANY OPTIONS TO IT
\usepackage{algorithm}
\usepackage{algorithmic}

\usepackage{newfloat}
\usepackage{listings}
\DeclareCaptionStyle{ruled}{labelfont=normalfont,labelsep=colon,strut=off} % DO NOT CHANGE THIS
\floatstyle{ruled}
\newfloat{listing}{tb}{lst}{}
\floatname{listing}{Listing}

\usepackage{booktabs}       % professional-quality tables
\usepackage{multirow}
\usepackage{graphicx}
\usepackage{bm}
\usepackage{amsmath,amsthm,amsfonts} 
\usepackage{enumerate}
\usepackage[table]{xcolor} % 引入xcolor宏包，并开启table选项
\usepackage{colortbl} % 引入colortbl宏包
\usepackage{xspace}
\usepackage{subcaption}

\usepackage[textsize=tiny]{todonotes}
\definecolor{coolgrey}{rgb}{0.55, 0.57, 0.67}

\newcommand{\algo}{\textsc{FtaT}\xspace}
\newcommand{\CDO}{\textsc{Cdo}\xspace}
\newcommand{\LCW}{\textsc{Lcw}\xspace}

\newcommand{\setting}{\emph{AdaTab}\xspace}

\title{Fully Test-time Adaptation for Tabular Data}
\author {
    Zhi Zhou\textsuperscript{\rm 1, \equalcontrib},
    Kun-Yang Yu\textsuperscript{\rm 1, \rm 2, \equalcontrib},
    Lan-Zhe Guo\textsuperscript{\rm 1, \rm 3, \thanks{Corresponding author}},
    Yu-Feng Li\textsuperscript{\rm 1, \rm 2, \textsuperscript{\textdagger}}
}
\affiliations {
    \textsuperscript{\rm 1}National Key Laboratory for Novel Software Technology, Nanjing University, China\\
    \textsuperscript{\rm 2}School of Artificial Intelligence, Nanjing University, China\\
    \textsuperscript{\rm 3}School of Intelligence Science and Technology, Nanjing University, China.\\
    \{zhouz, yuky, guolz, liyf\}@lamda.nju.edu.cn
}

\begin{document}
\maketitle

\begin{abstract}
    Tabular data plays a vital role in various real-world scenarios and finds extensive applications. Although recent deep tabular models have shown remarkable success, they still struggle to handle data distribution shifts, leading to performance degradation when testing distributions change. To remedy this, a robust tabular model must adapt to generalize to unknown distributions during testing. 
    In this paper, we investigate the problem of fully test-time adaptation (FTTA) for tabular data, where the model is adapted using only the testing data. We identify three key challenges: the existence of label and covariate distribution shifts, the lack of effective data augmentation, and the sensitivity of adaptation, which render existing FTTA methods ineffective for tabular data. 
    To this end, we propose the \emph{\textbf{F}ully \textbf{T}est-time \textbf{A}daptation for \textbf{T}abular data}, namely \algo, which enables FTTA methods to robustly optimize the label distribution of predictions, adapt to shifted covariate distributions, and suit a variety of tasks and models effectively.
    We conduct comprehensive experiments on six benchmark datasets, which are evaluated using three metrics. The experimental results demonstrate that \algo outperforms state-of-the-art methods by a margin. 
\end{abstract}

\section{Introduction}
Tabular data~\cite{naomi17tabular} plays a vital role in numerous practical applications, including economics~\cite{Salehpour2024economics}, healthcare~\cite{ching2018opportunities}, finance~\cite{ozbayoglu2020deep}, and manufacturing~\cite{HeinDTUHRS17}. 
Deep neural networks (DNNs) have recently shown remarkable success in handling tabular data, often surpassing traditional statistical methods when training and test data share the same distribution~\cite{Arik21TabNet, Gorishniy21FTTransformer}.
However, real-world applications often experience shifts in data distributions during testing, leading to significant performance degradation in existing methods~\cite{Kolesnikov23wildtab}.

To address distribution shifts during testing, fully test-time adaptation (FTTA) algorithms have emerged, enhancing the performance of pre-trained DNNs using only testing data.
These methods are particularly designed to deal with covariate distribution shift~\cite{WangSLOD21}, label distribution shift~\cite{WuGSW21}, or both~\cite{zhouz23ods}, adapting the model parameters~\cite{WangCoTTA22} or optimizing the predictions~\cite{BoudiafLAME22}. 
However, they are primarily designed for image tasks and heavily rely on image augmentation strategies~\cite{WangCoTTA22} and image-specific data assumptions~\cite{BoudiafLAME22, zhouz23ods}, rendering them less effective for tabular data. 
As a result, fully test-time adaptation for tabular data remains underexplored, despite its significance in real applications.

To this end, we study the fully test-time adaptation problem setting for tabular data, namely \setting, which holds significant practical value~\cite{naomi17tabular}. 
For example, in financial applications~\cite{kritzman2012regime}, the non-stationary financial market environment can cause significant changes in the data distribution between training and testing. For instance, shifts in the stock market can significantly affect market behavior and investor sentiment. 
These distribution shifts degrade the model performance and seriously affect investment decision-making and risk management, thereby leading to financial losses~\cite{GuoHY23}. 
The goal of the \setting problem setting is to adapt the trained deep tabular model to unknown distributions using only testing data, preventing the performance degradation caused by distribution shifts in downstream tabular applications. 

In this paper, we conduct an in-depth investigation into the \setting problem. 
Our four observations reveal three key challenges in designing FTTA methods for tabular data: 
(a) Covariate and label distribution shifts exist in tabular data, but they cannot be effectively addressed by existing FTTA methods; 
(b) Typical augmentation for test-time adaptation is often ineffective for tabular data, limiting the ability of FTTA methods to compute consistency; 
(c) Adaptation is sensitive to both tasks and models for tabular data.
To address these challenges, we propose a novel FTTA approach, \algo. It comprises three essential modules: \emph{Confident Distribution Optimizer}, \emph{Local Consistent Weighter}, and \emph{Dynamic Model Ensembler}, which robustly track and optimize the label distribution of predictions, adapt the model to shifted covariate distribution, and dynamically adapt the model for various tasks and models.
To summarize, the contributions of this paper are threefold:
\begin{enumerate}[(1)]
    \item We investigate fully test-time adaptation for tabular data, identifying three key challenges: the existence of label and covariate distribution shifts, the lack of effective data augmentation, and the sensitivity of adaptation.
    \item We propose a novel approach, \algo, which incorporates the \emph{Confident Distribution Optimizer}, \emph{Local Consistent Weighter}, and \emph{Dynamic Model Ensembler} to address the challenges of shifted label distribution, shifted covariate distribution, and sensitivity, respectively.
    \item We evaluate the \algo approach on six tabular benchmarks with real distribution shifts using three backbone models, demonstrating that the proposed approach significantly outperforms state-of-the-art FTTA methods.
\end{enumerate}

\section{Problem and Analysis}
In this section, we first introduce the \setting problem setting, including the notations and problem formulation. 
We then present four observations of \setting, which highlight three main challenges and underscore the necessity of designing FTTA methods specifically for tabular data.

\subsection{Problem Formulation}

We consider the fully test-time adaptation problem setting for tabular classification problem, namely, \setting. 
The input space is $\mathcal{X} \in \mathbb{R}^{d}$, where $d$ is the number of features. Each feature can be a continuous or discrete value.
The label space is $\mathcal{Y} \in \{0, 1\}^K$, where the $K$ is the number of classes. 
In this setting, we are given a well-trained source tabular model $f_{\theta_{0}}: \mathcal{X} \mapsto \mathcal{Y}$ with the initial parameters $\theta_0$. 
During the testing phase, the model is solely adapted based on the unlabeled batched testing data $D_t$ at each timestamp $t$, updating its parameters from $\theta_{t}$ to $\theta_{t+1}$. 
The goal of \setting problem is to adapt the initial given model $f_{\theta_{0}}$ during the testing phase, so that the adapted model $f_{\theta_{t}}$ can generalize better on the test data $D_{t}$ at each timestamp $t$. 
\begin{figure*}[t]
  \centering
  \begin{minipage}[t]{0.32\textwidth}
    \includegraphics[width=\linewidth]{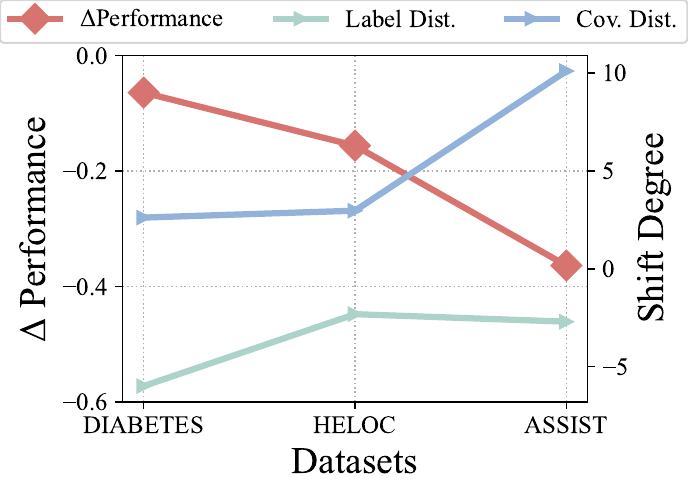}
    \caption{The label and covariate distribution shifts between training and testing in tabular data degrade the model performance. The shift degree is taken logarithm for aesthetic purposes.}
    \label{fig:shift}
  \end{minipage}
  \hfill
  \begin{minipage}[t]{0.265\textwidth}
    \includegraphics[width=\linewidth]{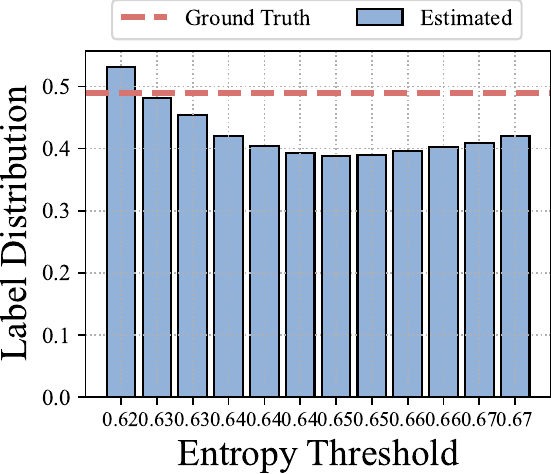}
    \caption{The label distribution of data for which the entropy of model predictions is lower than various thresholds. The ground truth is marked with dashed line.}
    \label{fig:lbs}
  \end{minipage}
  \hfill
  \begin{minipage}[t]{0.29\textwidth}
    \includegraphics[width=\linewidth]{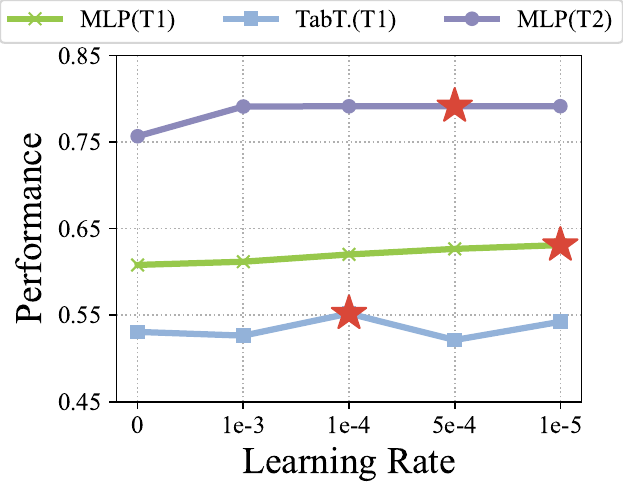}
    \caption{The performance of \algo with different learning rates. The optimal value differs across backbones and tasks. The highest point of each line is marked by a red star.}
    \label{fig:sensitive}
  \end{minipage}
\end{figure*}

\subsection{Problem Analysis}
\label{sec:analysis}

In the context of the \setting problem, we have identified four observations that also serve as key challenges hindering FTTA methods from effectively working with tabular data, in contrast to the standard fully test-time adaptation designed for image tasks.

\noindent \paragraph{\underline{Observation} 1: Covariate distribution and label distribution shifts in tabular data hinder performance of FTTA methods.}
Our first observation reveals that both covariate distribution and label distribution shifts exist in tabular data, and both contribute to performance degradation. 
To estimate the distribution shifts between the training and testing datasets, we use the optimal transport dataset distance with Gaussian approximation~\cite{alvarez2020geometric} to measure covariate distribution shifts and the $L_2$ distance~\cite{GardnerPS23} to assess label distribution shifts. 
As shown in Fig.~\ref{fig:shift}, both increases in label distribution shift (DIABETE$\rightarrow$HELOC) and covariate distribution shift (HELOC$\rightarrow$ASSIST) degrade performance on the testing data with distribution shifts. 
However, as our experimental results reveal, existing robust FTTA methods, such as ODS~\cite{zhouz23ods}, designed to address covariate and label distribution shifts in tabular data, do not perform well in practice.
This observation highlights the challenges faced by FTTA methods designed for tabular data in addressing both covariate and label distribution shifts simultaneously. 
We also estimate the label distribution of data whose model prediction entropy is lower than thresholds on DIABETES dataset. The results in Fig.~\ref{fig:lbs} demonstrate that we can still accurately estimate the label distribution using only low-entropy data. 

\noindent \paragraph{\underline{Observation} 2: Typical Augmentation used in test-time adaptation is ineffective for tabular data.}

Existing FTTA methods, such as CoTTA~\cite{WangCoTTA22} and AdaContrast~\cite{chen19adacontrast}, rely heavily on data augmentation. However, augmentation for tabular data is not as effective as it is for images. 
We conduct experiments based on CoTTA methods with perturb augmentation~\cite{FangTCZLZZ22} with different perturbation strengths controlled by $\sigma$. 
As shown in Table~\ref{tab:obs2}, the performance of CoTTA degrades as the augmentation strength increases, and it fails to surpasses the non-adaptation baseline. This observation highlights the challenges of FTTA methods designed for tabular data, particularly their inability to rely on data augmentation method when dealing with tabular data.

\begin{table}[t]
  \caption{Performance of the non-adaptation baseline and CoTTA method with different augmentation strengths $\sigma$ using the MLP model. The best performance is in \textbf{bold}.}
  \label{tab:obs2}
  \centering
  \resizebox{\linewidth}{!}{
  \begin{tabular}{l|ccc}
  \bottomrule
  \toprule
  Method & DIABETE & HELOC & ASSIST \\
  \midrule
  Non-Adaptation & \bm{$60.82\pm0.22$} & \bm{$54.37\pm5.35$} & \bm{$55.86\pm3.81$} \\
  $\sigma$ = 0.2 & $60.46\pm0.20$ & $46.40\pm3.08$ & $54.89\pm1.88$ \\
  $\sigma$ = 0.4 & $59.18\pm0.42$ & $43.36\pm0.25$ & $54.86\pm3.00$ \\
  $\sigma$ = 0.6 & $57.73\pm0.64$ & $43.06\pm0.07$ & $54.51\pm2.26$ \\
  $\sigma$ = 0.8 & $56.19\pm0.83$ & $43.07\pm0.03$ & $53.79\pm3.80$ \\
  $\sigma$ = 1.0 & $54.74\pm0.77$ & $43.09\pm0.01$ & $54.23\pm3.56$ \\
  \bottomrule
  \toprule
  \end{tabular}}
\end{table}

\noindent \paragraph{\underline{Observation} 3: Adaptation is sensitive to both tasks and models for tabular data. }

Unlike images, which exhibit strong transferability~\cite{HeGD19} and similar structure~\cite{torralba2003statistics}, tabular data from different tasks differs significantly. Moreover, different backbone models~\cite{Gorishniy21FTTransformer, tabtransformer} are designed to address distinct tasks. 
Therefore, for specific tabular tasks and the backbone models used, FTTA methods require tuning for optimal performance.
As shown in Fig.~\ref{fig:sensitive}, the optimal learning rates for different backbone models on the same task and same backbone model on different tasks varies. 
This observation indicates that the \setting problem requires a model capable of dynamically tuning the learning rates, rather than relying on a fixed learning rate.

\noindent \paragraph{\underline{Observation} 4: Existing FTTA methods degradates when dealing with tabular data.}
Observations 1, 2, and 3 also serve as key challenges in designing FTTA methods for tabular data, causing existing FTTA methods to fail to improve performance compared to the baseline. 
As shown in Tab.~\ref{tab:obs4}, we compare performance of non-adaptation baseline with two representative FTTA methods which respectively optimize model parameters and predictions, i.e., TENT~\cite{WangSLOD21} and LAME~\cite{BoudiafLAME22}. 
As the covariate distribution and label distribution shifts become more severe (DIABETE$\rightarrow$HELOC$\rightarrow$ASSIST), both types of FTTA methods fail to surpass the non-adaptation baseline. 
This observation highlights the significance of developing FTTA methods specifically designed for tabular data which addressing three challenges simultaneously.

\begin{table}[t]
  \caption{Performance of the non-adaptation baseline and two representative FTTA methods using an MLP backbone model. Degraded performance is \underline{underlined}.}
  \label{tab:obs4}
  \centering
  \resizebox{\linewidth}{!}{
  \begin{tabular}{l|ccc}
  \bottomrule
  \toprule
  Method & DIABETE & HELOC & ASSIST \\
  \midrule
  Non-Adaptation       & $60.82\pm0.22$ & $54.37\pm5.35$             & $55.86\pm3.81$ \\
  Optimize Parameters  & $61.34\pm0.33$ & \underline{$54.35\pm5.38$} & \underline{$50.87\pm0.32$} \\
  Optimize Predictions & $61.47\pm0.35$ & \underline{$43.10\pm0.00$} & \underline{$45.12\pm0.18$} \\
  \bottomrule
  \toprule
  \end{tabular}}
\end{table}

\section{Methodology}

In this section, we introduce our \algo approach for \setting problem setting. 
As discussed in the analysis section, the \setting problem encompasses three challenges:
\begin{enumerate}
    \item[(a)] Covariate and label distribution shifts exist in tabular data, but cannot be effectively addressed by existing FTTA methods; 
    \item[(b)] Typical augmentation used for test-time adaptation is not very effective for tabular data, limiting the ability of FTTA methods to adopt consistency; 
    \item[(c)] Adaptation is sensitive to both tasks and models for tabular data.
\end{enumerate}

To address the above challenges, we introduce three modules specifically designed for \setting problem, i.e., \emph{Confident Distribution Optimizer}, \emph{Local Consistent Weighter}, and \emph{Dynamic Model Ensembler}. 
Specifically, the \emph{Confident Distribution Optimizer} optimizes the original model predictions $f_{\theta_t}(x)$ to $\widehat{f}_{\theta_t}(x)$ for a data point $x$ at timestamp $t$.
The \emph{Local Consistent Weighter} affects the adaptation objective:
\begin{equation}
    \theta_{t+1} = \mathop{\arg\min}\limits_{\theta} \sum_{i=1}^{D_t} \mathcal{W}(x_i, D_t, \theta_t) \cdot \text{Loss} \left (\widehat{f}_{\theta_{t}}(x_i) \right )
\end{equation}
using a weighting function $\mathcal{W}$, where $\text{Loss}(\cdot)$ represents the unsupervised loss for test-time adaptation, and we employ entropy loss in accordance with classical methods.
The \emph{Dynamic Model Ensembler} maintains multiple models and ensembles their predictions in an online manner. We will introduce them in detail.

\subsection{Confident Distribution Optimizer} 

First, we aim to optimize the model predictions to align with the current shifted label distribution. The existing solution~\cite{zhouz23ods} fails because the challenging nature of tabular data prevents the model from making accurate predictions, which in turn hinders the estimation of the label distribution. Therefore, the key challenge is how to robustly track the shifted label distribution $\widehat{P}_t$ at each timestamp $t$.
With original label distribution $P_0$ and esitmated label distribution $\widehat{P}_t$, the optimized model prediction $\widehat{f}_{\theta_{t+1}}(x_k)$ for next timestamp on data point $x_k$ is
\begin{equation}
    \widehat{f}_{\theta_{t+1}}(x_k) = \frac{f_{\theta_{t+1}}(x_k) \circ \widehat{P}_t}{P_0}
\end{equation}

Motivated by our observations, we recognize that we can estimate the label distribution $\widetilde{P}_t$ with bias from model $\widehat{f}_{\theta_t}$ at each timestamp $t$ using only data with low-entropy predictions (i.e., data with confident predictions): 
\begin{equation}
    \widetilde{P}_t = \frac{ \sum_{i=1}^{|D_t|} \mathbb{I} \left [ \text{Entropy} \left (\widehat{f}_{\theta_t}(x_i) \right ) < \epsilon \right ] \cdot \widehat{f}_{\theta_t}(x_i) }{\sum_{i=1}^{|D_t|} \mathbb{I} \left [ \text{Entropy} \left (\widehat{f}_{\theta_t}(x_i) \right ) < \epsilon \right ]}
\end{equation}
where $D_t$ is current data batch, $\epsilon$ is a threshold, and Entropy($\cdot$) is the function for computing entropy of predictions. Note that there exists bias in esitmated $\widetilde{P}_t$ as the model predictions may contains errors. To address this issue, we compute the covariate matrix $\widehat{C}_t$ at the current timestamp $t$, where its $k$-th row is equal to
\begin{equation}
    \frac{\sum_{i=1}^{|D_t|}  \mathbb{I} \left [ \mathop{\arg\max}_j \widehat{f}_{\theta_t}(x_i)_j = k\right ] \cdot \widehat{f}_{\theta_t}(x_i)}{\sum_{i=1}^{|D_t|}  \mathbb{I} \left [ \mathop{\arg\max}_j \widehat{f}_{\theta_t}(x_i)_j = k\right ]}
\end{equation}
Then, the unbiased label distribution is $\widehat{C}_t^{-1} \widetilde{P}_t$. 
We additionally adopt a temperal ensemble method~\cite{LaineA17} to robustly track estimated label distribution smoothly with a factor $\alpha$ and previous estimated shift $\widehat{P}_{t-1}$: 
\begin{equation}
    \widehat{P}_t = \text{Norm} \left (\widehat{P}_{t-1} - \alpha \cdot \widehat{C}_t^{-1} \widetilde{P}_t \right )
\end{equation}
where $\text{Norm}(\cdot)$ normalizes the distribution to sum to one, and we use the Softmax function for this purpose.

\begin{figure}[t]
    \centering
    \includegraphics[width=\linewidth]{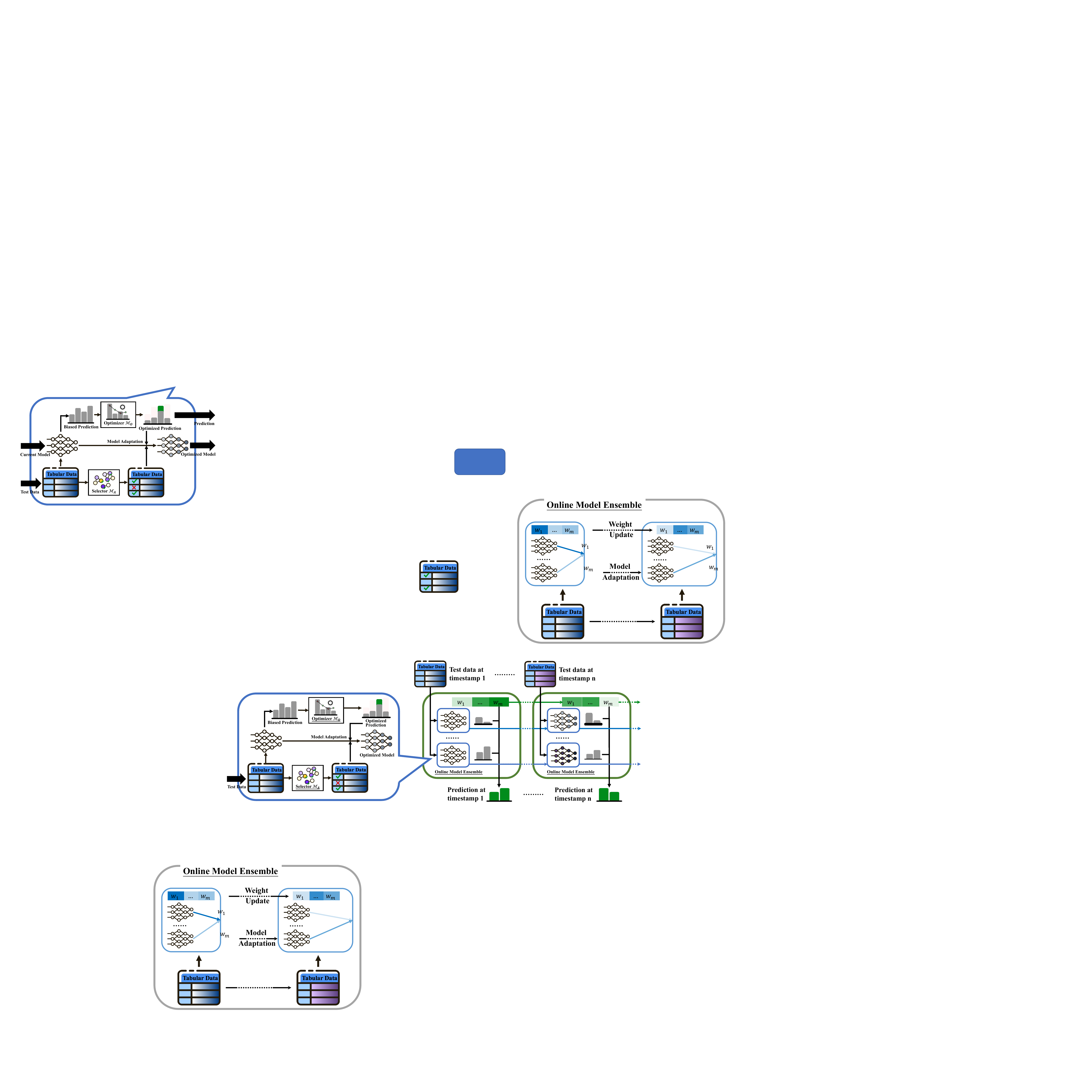}
    \caption{The overall illustation of \algo approach. }
    \label{fig:framework}
\end{figure}

\subsection{Local Consistent Weighter}

Second, to mitigate the adverse effects of shifted covariate distribution, we propose filtering testing data with low-quality predictions to ensure robust test-time adaptation and avoiding error accumulation. However, for tabular data, computing consistency through data augmentation is non-trivial because our observation indicates that augmentation for tabular data is not as reliable as it is for image data. 

To address this issue, we propose replacing the consistency between a data point and its augmentations with the consistency between a data point and its neighborhood, under the inspiration of one existing tablar study~\cite{GorishniyRKSKB24}.
Specifically, we define the neighborhood set $N(x_k, D_t)$ of each data point $x_k$ in current batch $D_t$ measured by one distance function $\text{Dist}(\cdot, \cdot)$:
\begin{equation}
    N(x_k, D_t) = \left \{x| \text{Dist}(x, x_k) < \overline{\text{Dist}_t}, x \in D_t \right \}
\end{equation} 
where $\overline{\text{Dist}_t} = \frac{2}{|D_t| (|D_t|-1)}\sum_{i=1}^{|D_t|} \sum_{j=i+1}^{|D_t|} \text{Dist}(x_i, x_j)$ is the average pair-wise distance in $D_t$ and we adopt $L_2$ distance as the distance function. 
Next, we define the prediction of one data point $x_k$ is consistent if its soft pseudo-label vector is close to the average soft pseudo-label vectors in its neighborhood $N(x_k, D_t)$. 
Then, we define the indication function $\mathcal{I}(x_k, D_t, \theta_t)$ to decide whether one data point $x_k$ in current batch $D_t$ is consistent: 
\begin{equation}
    \label{eq:indicator}
    \mathcal{I}(x_k, D_t, \theta_t) = 
    \begin{cases}
        1, &\left \|f_{\theta_t}(x_k) - \frac{\sum_{x \in N(x_k, D_t)} f_{\theta_t}(x)}{|N(x_k, D_t)|} \right \| < \beta, \\
        0, &Otherwise.
    \end{cases}
\end{equation}
where $\theta_t$ is the parameters of model at timestamp $t$, $f_{\theta_t}(\cdot)$ predicts pseudo-label of data point, and $\beta$ is a hyperparameter to control the degree of consistency. 
To additionally ensure the robustness of adaptation, we compute the uncertainty of each data point using margin of prediction~\cite{HeltonJ11} as $\max \hat{f}_{\theta_t}(x_k) - \min \hat{f}_{\theta_t}(x_k)$. 
Finally, our proposed local consistent weighter $\mathcal{W}(x_k, D_t, \theta_t)$ is formulated as follows: 
\begin{equation}
    \left [\max \hat{f}_{\theta_t}(x_k) - \min \hat{f}_{\theta_t}(x_k) \right ] \cdot \mathcal{I}(x_k, D_t, \theta_t)
\end{equation}

\subsection{Dynamic Model Ensembler} 

Third, to address the sensitivity issue of adaptation, we employ the online ensemble learning paradigm~\cite{bai2022adapting} to optimize multiple models with different learning rates and ensemble their outputs through weighted averaging to obtain the overall robust prediction. 
Specifically, we maintain $M$ models using different learning rates during the testing, denoted as the set of base models $\left \{\widehat{f}_{\theta_t^i} \right \}_{i=1}^{M}$.
Then, model predictions are weighted according to corresponding loss values $w_i \propto 1 - R^i_t(D_t)$, where $R^i_t(D_t)$ is the loss value of $i$-th model $\widehat{f}_{\theta_t^i}$ evaluated on current batched data $D_t$ and satisfies the constraint $\sum_{i=1}^M w_i = 1$. 
The final prediction of the \algo approach is obtained by weighted ensemble for a data point $x$, that is, $\sum_{i=1}^M w_i \cdot \widehat{f}_{\theta_t^i}(x)$.

\section{Experiments}
\label{sec:experiments}
In this section, we first introduce the experimental setup.
Next, we present our empirical results, comparing our \algo approach with existing FTTA methods across four benchmark datasets.
Finally, we conduct an ablation study and provide further analysis for our proposed method.

\begin{table*}[t]
    \caption{The average performance of \algo approach and comparison methods using three backbone models. The best performance is in \textbf{bold}. Our \algo approach achieves the best performance across all three backbone models.}
    \label{tab:exps-average}
    \centering
    \resizebox{\linewidth}{!}{
    \begin{tabular}{l|ccc|ccc|ccc}
    \bottomrule
    \toprule
    \multirow{2}{*}{Method} & \multicolumn{3}{c|}{MLP} & \multicolumn{3}{c|}{TabTransformer} &\multicolumn{3}{c}{FT-Transformer}  \\ \cmidrule{2-10} 
      & Acc. & Balanced Acc. & F1  & Acc. & Balanced Acc. & F1 & Acc. & Balanced Acc. & F1   \\
    \midrule
    Non-Adaptation	 & $62.45$ & $64.61$ & $60.59$ & $60.86$ & $63.08$ & $58.32$ & $59.69$ & $62.49$ & $54.29$\\
    TENT	         & $58.43$ & $61.63$ & $50.97$ & $58.32$ & $61.40$ & $51.73$ & $51.61$ & $55.41$ & $36.34$\\
    EATA	         & $61.43$ & $63.69$ & $60.11$ & $60.33$ & $62.36$ & $60.09$ & $56.04$ & $58.95$ & $44.62$\\
    LAME	         & $59.48$ & $62.32$ & $58.47$ & $59.15$ & $62.50$ & $58.39$ & $58.90$ & $61.98$ & $51.86$\\
    CoTTA	         & $61.59$ & $63.78$ & $60.57$ & $60.37$ & $62.82$ & $59.75$ & $59.64$ & $62.43$ & $53.41$\\
    ODS	             & $59.18$ & $62.15$ & $57.83$ & $59.22$ & $62.02$ & $58.46$ & $59.05$ & $61.70$ & $51.41$\\
    SAR	             & $61.16$ & $63.49$ & $59.18$ & $60.30$ & $62.77$ & $59.72$ & $59.27$ & $62.11$ & $57.04$\\
    \hline
    \rowcolor{gray!30}\algo	 & \bm{$66.77$} & \bm{$64.96$} & \bm{$72.00$} & \bm{$66.14$} & \bm{$64.40$} &\bm{$69.03$} & \bm{$64.01$} & \bm{$62.54$} &\bm{$69.56$}\\
    \bottomrule
    \toprule
    \end{tabular}}
\end{table*}

\subsection{Experimental Setup}

\paragraph{Evaluation Protocol.} 
In our experiments on tabular tasks, we follow the fully test-time adaptation setting, where the source model is trained on training data and adapted to shifted test data without any access to the source training data. 
Specifically, we train the source model on training data and select the best model based on the validation set following the TableShift benchmark~\cite{GardnerPS23}. 
Then, \algo approach and existing FTTA methods are evaluated on the shifted test set. 
We select six common tabular benchmark datasets from the TableShift benchmark, which exhibit significant performance gaps under distribution shifts.
Therfore, these datasets contains samples from 10K to 5M and features from 26 to 365, which can cover a wide range of tabular scenarios under distribution shifts.
All experiments are repeated with different random seeds, and the mean and standard deviation are reported.

\paragraph{Comparison Methods.} 
To compare our \algo apporach with various FTTA methods, including typical FTTA methods (i.e., TENT~\cite{WangSLOD21} and EATA~\cite{NiuW0CZZT22}), continual FTTA methods (i.e., CoTTA~\cite{WangCoTTA22}), and recently proposed robust FTTA methods (i.e., SAR~\cite{niu2023towards}, LAME~\cite{BoudiafLAME22}, and ODS~\cite{zhouz23ods}).

\subsection{Main Results}

To evaluate the effectiveness of \algo, we report the detailed experimental results using three backbone models in Tab.~\ref{tab:exps-average}. 
The performance are measured by three metrics including accuracy, balanced accuracy, and F1 score. 
The experimental results show our \algo approach outperforms existing methods by a margin on all metrics. 

Moreover, we report the detailed results on each dataset using MLP backbone model in Tab.~\ref{tab:exps-mlp}. 
\algo achieves the best performance on major cases and give competitive performance on the resting cases, demonstrating the effectiveness of \algo on various tabular datasets with different backbone models. ODS~\cite{zhouz23ods} addresses both the covariate and label distribution shifts, however, it underperforms \algo on most cases, demonstrating the effectiveness of \algo in handling distribution shifts for tabular data. 
The detailed results using FT-Transformer~\cite{Gorishniy21FTTransformer} and TabTransformer~\cite{tabtransformer} are included in our supplementary material due to the space limits. 

\begin{table*}[t]
    \caption{Performance of \algo approach and comparison methods on 6 datasets using MLP. The best performance is in \textbf{bold}.}
    \label{tab:exps-mlp}
    \centering
    \resizebox{\linewidth}{!}{
    \begin{tabular}{l|ccc|ccc|ccc}
    \bottomrule
    \toprule
    \multirow{2}{*}{Method} & \multicolumn{3}{c|}{HELOC} & \multicolumn{3}{c|}{ANES} &\multicolumn{3}{c}{Health Ins.}  \\ \cmidrule{2-10} 
      & Acc. & Balanced Acc. & F1  & Acc. & Balanced Acc. & F1 & Acc. & Balanced Acc. & F1   \\
    \midrule
    Non-Adaptation	 & $54.37\pm5.35$	 & $58.25\pm3.56$	 & $40.02\pm16.8$	 & $79.11\pm0.31$	 & $75.66\pm0.46$	 & \bm{$84.24\pm0.16$}	& $65.79\pm0.63$ & \bm{$70.68\pm0.44$} & $66.21\pm0.90$\\
    TENT	 & $54.35\pm5.38$	 & $58.24\pm3.58$	 & $39.95\pm16.9$	 & $78.07\pm0.35$	 & $74.09\pm0.65$	 & $83.76\pm0.13$	& $64.30\pm0.70$ & $69.79\pm0.47$ & $63.87\pm1.06$\\
    EATA	 & $54.37\pm5.35$	 & $58.25\pm3.56$	 & $40.02\pm16.8$	 & $78.13\pm0.30$	 & $74.20\pm0.59$	 & $83.79\pm0.10$	& $65.78\pm0.63$ & $70.68\pm0.44$ & $66.21\pm0.90$\\
    LAME	 & $43.10\pm0.00$	 & $50.00\pm0.00$	 & $30.10\pm0.00$	 & $63.50\pm0.00$	 & $54.60\pm0.00$	 & $46.80\pm0.00$	& $63.44\pm1.69$ & $69.14\pm1.09$ & $62.61\pm2.69$\\
    CoTTA	 & $54.36\pm5.35$	 & $58.25\pm3.56$	 & $40.03\pm16.8$	 & $78.13\pm0.30$	 & $74.20\pm0.59$	 & $83.79\pm0.10$	& $65.79\pm0.63$ & \bm{$70.68\pm0.44$} & $66.21\pm0.90$\\
    ODS	 & $43.10\pm0.00$	 & $50.00\pm0.00$	 & $30.10\pm0.00$	 & $63.50\pm0.00$	 & $54.60\pm0.00$	 & $46.80\pm0.00$	& $63.45\pm1.68$ & $69.14\pm1.07$ & $62.62\pm2.68$  \\
    SAR	 & $52.32\pm6.05$	 & $56.74\pm3.99$	 & $33.16\pm19.0$	 & $78.13\pm0.30$	 & $74.20\pm0.59$	 & $83.79\pm0.10$	& $65.79\pm0.63$ & \bm{$70.68\pm0.44$} & $66.21\pm0.90$\\
    \hline
    \rowcolor{gray!30}\algo	 & \bm{$64.09\pm1.14$}	 & \bm{$63.64\pm0.93$}	 & \bm{$67.80\pm2.71$}	 & \bm{$80.09\pm0.23$}	 & \bm{$79.12\pm0.20$}	 & $83.42\pm0.25$	& \bm{$72.42\pm0.20$} & $65.30\pm0.15$ & \bm{$80.83\pm0.23$}\\
    \midrule \midrule
    \multirow{2}{*}{Method} & \multicolumn{3}{c|}{ASSIST} & \multicolumn{3}{c|}{DIABETE} &\multicolumn{3}{c}{Hypertension} \\ \cmidrule{2-10} 
      & Acc. & Balanced Acc. & F1  & Acc. & Balanced Acc. & F1  & Acc. & Balanced Acc. & F1  \\
    \midrule
    Non-Adaptation	 & $55.86\pm3.81$	 & $60.81\pm3.37$	 & $66.42\pm1.86$	 & $60.81\pm0.21$	 & $60.59\pm0.24$	 & $51.18\pm1.69$	& $58.76\pm1.68$ & \bm{$61.69\pm0.95$} & $55.46\pm4.03$\\
    TENT	 & $50.87\pm0.32$	 & $56.41\pm0.29$	 & $63.99\pm0.15$	 & $61.34\pm0.33$	 & $61.15\pm0.34$	 & $53.75\pm1.01$	& $41.67\pm0.08$ & $50.07\pm0.05$ & $0.49\pm0.36$\\
    EATA	 & $55.86\pm0.18$	 & $60.81\pm0.16$	 & $66.42\pm0.08$	 & $61.36\pm0.30$	 & $61.16\pm0.31$	 & $53.68\pm1.09$	& $57.81\pm2.32$ & $61.19\pm1.38$ & $52.87\pm5.82$\\
    LAME	 & $45.12\pm0.18$	 & $51.30\pm0.18$	 & $61.40\pm0.18$	 & $61.47\pm0.35$	 & $61.30\pm0.37$	 & $54.67\pm1.45$	& $58.63\pm1.60$ & $61.64\pm0.92$ & $55.12\pm3.84$\\
    CoTTA	 & $55.86\pm0.18$	 & $60.81\pm0.16$	 & $66.42\pm0.08$	 & $61.39\pm0.29$	 & $61.20\pm0.30$	 & $53.82\pm1.05$	& $58.76\pm1.68$ & \bm{$61.69\pm0.95$} & $55.46\pm4.03$\\
    ODS	 & $45.12\pm0.18$	 & $51.30\pm0.18$	 & $61.40\pm0.18$	 & $61.47\pm0.35$	 & $61.30\pm0.37$	 & $54.69\pm1.43$	& $57.12\pm1.46$ & $60.80\pm0.93$ & $51.41\pm3.43$\\
    SAR	 & $55.86\pm0.18$	 & $60.81\pm0.16$	 & $66.42\pm0.08$	 & $61.38\pm0.30$	 & $61.19\pm0.30$	 & $53.98\pm0.93$	& $58.21\pm1.51$ & $61.50\pm0.77$ & $53.81\pm4.05$\\
    \hline
    \rowcolor{gray!30}\algo	 & \bm{$60.17\pm2.87$}	 & \bm{$63.79\pm2.18$}	 & \bm{$66.92\pm1.20$}	 & \bm{$61.66\pm0.30$}	 & \bm{$61.54\pm0.28$}	 & \bm{$59.27\pm0.96$}	& \bm{$62.20\pm0.94$} & $56.36\pm1.62$ & \bm{$73.77\pm0.13$}\\
    \bottomrule
    \toprule
    \end{tabular}}
\end{table*}

Our experimental results confirm our last observation, showing that existing FTTA methods face performance degradation on tabular data with distribution shifts, thereby demonstrating the necessity to study the \setting problem. Our \algo approach consistently outperforms non-adaptation baseline and existing FTTA methods in most cases, offering insights into this challenging problem.

\subsection{Further Analysis}

\paragraph{Ablation Study. }

We analyze the effectiveness of the \emph{Confident Distribution Optimizer} (denoted as \CDO) and the \emph{Local Consistent Weighter} (denoted as \LCW) in Tab.~\ref{tab:ablation-mlp} on DIABETE and HELOC datasets using the MLP backbone model. 
The experimental results of \algo apporach without \CDO and \LCW are reported measured by three metrics. 
Without \CDO, the performance of \algo approach improves marginally, which indicates that the label distribution shift hinders the performance and \CDO addressing the label distribution shift plays a more important role in the \algo approach. 
Without \LCW, the performance of \algo cannot achieve the optimal level, demonstrating the essential role of \LCW to robustly update the model.
Overall, our \algo approach achieves the best performance when both \CDO and \LCW are employed, demonstrating their effectiveness in addressing challenges of FTTA for tabular data. 

\begin{table}[t]
    \caption{Ablation study of \algo approach on DIABETE dataset using MLP backbone model. The best performance is in \textbf{bold}. The results show that both \emph{Confident Distribution Optimizer} (denoted as \CDO) and the \emph{Local Consistent Weighter} (denoted as \LCW) are essential for \algo approach. }
    \label{tab:ablation-mlp}
    \centering
    \resizebox{\linewidth}{!}{
    \begin{tabular}{l|ccc}
   \bottomrule
    \toprule
    \multicolumn{4}{c}{\bf DIABETE} \\
    \hline
    Method & Acc. & Balanced Acc. & F1 \\
    \midrule
    Non-Adaptation	          & $60.81\pm0.21$	      & $60.59\pm0.23$	      & $51.18\pm1.69$        \\
    \algo w/o \CDO          & $60.85\pm0.22$	      & $60.61\pm0.24$	      & $51.26\pm1.69$       \\
    \algo w/o \LCW         & $61.43\pm0.16$        &$61.28\pm0.20$         &$55.61\pm1.67$         \\ \hline
    \rowcolor{gray!30}\algo                       & \bm{$61.66\pm0.30$}	  & \bm{$61.54\pm0.28$}	  & \bm{$59.27\pm0.96$}	\\
    \bottomrule
    \toprule
    \multicolumn{4}{c}{\bf HELOC} \\
    \hline
    Method & Acc. & Balanced Acc. & F1 \\
    \midrule
    Non-Adaptation	        & $54.37\pm5.35$	      & $58.25\pm3.56$	      & $40.02\pm16.8$        \\
    \algo w/o \CDO          & $62.73\pm0.23$ & $62.55\pm0.52$ & $66.45\pm0.91$ \\
    \algo w/o \LCW          & $62.52\pm0.86	$ &$61.73\pm1.06$  &$67.07\pm1.20$ \\ \hline
    \rowcolor{gray!30}\algo & \bm{$64.09\pm1.14$}	  & \bm{$63.64\pm0.93$}	  & \bm{$67.80\pm2.71$}	\\
    \bottomrule
    \toprule
    \end{tabular}}
    \vspace{-0.6em}
\end{table}

\begin{figure}[t]
    \centering
    \includegraphics[width=0.8\linewidth]{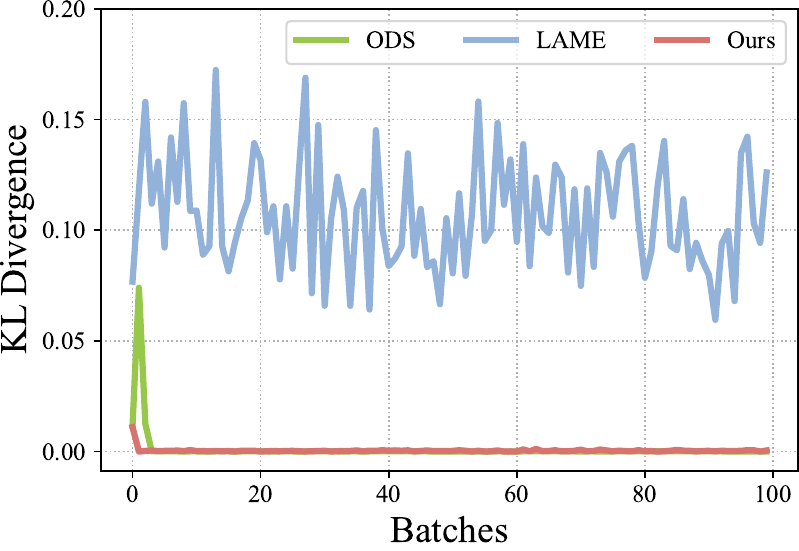}
    \caption{The performance of LAME, ODS, and \algo in estimating label distribution evaluated using KL divergence.}
    \label{fig:kldiv}
    \vspace{-0.7em}
\end{figure}

\paragraph{Estimation of Label Distribution. }
We compare the performance in estimating label distribution on the DIABETE dataset using the MLP backbone model. We adopt KL divergence to measure the distance between the ground-truth label distribution and its estimation. As shown in Fig.~\ref{fig:kldiv}, the LAME method cannot accurately estimate the label distribution. While the ODS method can robustly track the label distribution, it requires several iterations to converge to an accurate estimation. In contrast, our \algo approach achieves accurate label distribution estimation at a much faster speed. This result demonstrates the superiority of \algo approach.

\paragraph{Effects of Dynamic Model Ensembler. }
To validate the effectiveness of \emph{Dynamic Model Ensembler}, we conduct experiments running with base models with different learning rates, ensemble baseline, and \algo approach. Here, we compare with four base models with different learning rates $\{1e-3, 1e-4, 5e-4, 1e-5\}$. The ensemble baseline are the direct ensemble of these base models. 
\ref{tab:ense-study-mlp} presents the average performance using MLP model on DIABETE dataset. 
The results show that our \emph{Dynamic Model Ensembler} module consistently outperforms the average ensemble baseline, demonstrating the its effectiveness. Moreover, the \algo approach can achieve the best performance or competitive performance when compared to base learners with the optimal learning rate without requiring tuning, indicating the advantage of our \emph{Dynamic Model Ensembler} module.

\begin{table}[t]
    \caption{Performance of base models with different learning rates, directly ensemble of base models, and the \algo approach on DIABETE dataset using the MLP backbone model. The best performance is in \textbf{bold}. }
    \label{tab:ense-study-mlp}
    \centering
    \resizebox{\linewidth}{!}{
    \begin{tabular}{l|ccc}
    \bottomrule
    \toprule 
    Method & Acc. & Balanced Acc. & F1 \\
    \midrule
    Lr=1e-3    &  $61.49\pm0.46$	& $61.47\pm0.42$	&\bm{$60.39\pm0.73$}	\\
    Lr=1e-4    &  $61.58\pm0.30$	& $61.53\pm0.26$	&$59.37\pm0.95$ 	\\
    Lr=5e-4	   &  $61.56\pm0.33$	& $61.53\pm0.30$	&$59.93\pm0.90$\\
    Lr=1e-5    &  $61.58\pm0.30$	& $61.52\pm0.26$	&$59.23\pm0.97$ \\
    Average Ensemble &	 $61.57\pm0.31$	& $61.51\pm0.27$	&$59.24\pm0.95$	\\ \hline
    \rowcolor{gray!30}\algo      &	\bm{$61.60\pm0.31$}	& \bm{$61.53\pm0.27$}	&$59.27\pm0.96$	\\
    \bottomrule
    \toprule
    \end{tabular}}
    \vspace{-0.6em}
\end{table}

\begin{figure*}[t]
    \centering
    \includegraphics[width=0.9\linewidth]{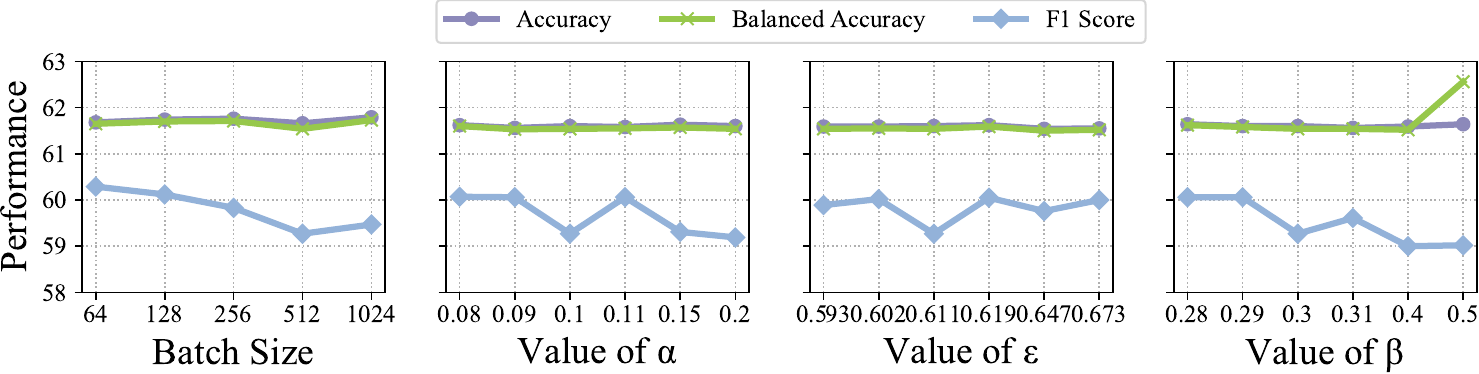}
    \caption{Robustness of batch size and hyperparameters $\alpha, \epsilon, \beta$ on DIABETE dataset using MLP backbone model. The results indicate that minor perturbations to the hyperparameters of \algo do not significantly affect its performance, demonstrating the practical robustness of \algo.}
    \label{fig:hyper}
    \vspace{-0.7em}
\end{figure*}

\paragraph{Robustness of Batch Size. }
In the main experiments, the batch size of the data stream is set to 512 due to the large quantity of testing data.
A natural question that arises is how the batch size affects the performance of the proposed method.
We conduct experiments on the DIABETE dataset using an MLP backbone model, with batch sizes set to $\{64, 128, 256, 512, 1024\}$.
As shown in Fig.~\ref{fig:hyper}, the results indicate that the accuracy and balanced accuracy metrics are robust across different batch sizes. Regarding the F1 score, it decreases as the batch size increases. 
Nevertheless, \algo consistently outperforms existing methods by a margin.

\paragraph{Robustness of Hyperparameters. }

To validate whether our proposed \algo approach is robust to the choices of hyperparameters, we conduct hyperparameter robustness experiments on DIABETE dataset evaluated by three metrics using MLP backbone model. 
Specifically, \algo contains three hyperparameters, i.e., $\epsilon$, $\alpha$ and $\beta$.
The hyperparameter $\alpha$ controls the rate at which the estimated label distribution is updated, enhancing the robustness of the \algo to estimation errors in certain batches.
The hyperparameter $\epsilon$ governs the entropy-based confident samples selection to accurately estimate the label distribution. 
$\beta$ determines the construction of the neighbor set for entropy minimization, contributing to robust model adaptation. 
We conducted three runs of experiments for each set of hyperparameters with $\alpha$ in $\{0.08, 0.09, 0.10, 0.11, 0.15, 0.20\}$, $\epsilon=\text{Entropy}([p, 1-p])$ where $p$ was set to $\{0.72, 0.71, 0.70, 0.69, 0.65, 0.60\}$, and $\beta$ in $\{0.28, 0.29, 0.30, 0.31, 0.40, 0.50\}$.
Fig.~\ref{fig:hyper} reports the average performance of each hyperparameter evaluated using three metrics on DIABETE dataset.
The results demonstrate that \algo is robust to slight changes in all hyperparameters.

\section{Related Work}
In this section, we mainly discuss two lines of related work, including test-time adaptation and deep tabular learning. 

\noindent \paragraph{Test-time Adaptation.} 
Test-time adaptation aims to adapt a source model to the distribution shift in testing data without using any source data. Previously, test-time training studies, such as TTT~\cite{SunWLMEH20} and TTT+~\cite{LiuKDBMA21}, manipulated the model in both the training and testing phases. 
They introduce self-supervised objectives at training time and adapt the model parameters by optimizing self-supervised objectives at testing time. However, when training data is inaccessible and model training cannot be controlled, test-time training paradigms become ineffective. 
Fully test-time adaptation aims to tackle this limitation by adapting the model without assumptions on the source model. Tent~\cite{WangSLOD21} updates the parameters of the BN layer at test time. EATA~\cite{NiuW0CZZT22} additionally conducts active sample selection and weighting strategies for efficiency. Other studies~\cite{gong2022note,goyal2022test} also propose diverse methods to adapt the BN layer to the test data distribution to ensure performance. 
In practice, SAR~\cite{niu2023towards} introduces a flat minimum optimization method to ensure generalization performance when the test batch size varies. CoTTA~\cite{WangCoTTA22} works on continually non-i.i.d. scenarios using weight-averaged models, augmentation-averaged predictions, and stochastically restoring. LAME~\cite{BoudiafLAME22} proposes a conservative approach to revise the model’s predictions instead of model parameters. ODS~\cite{zhouz23ods} focuses on test-time adaptation settings where covariate and label distributions change together. 
Recent TTA studies mainly focus on images and natural language, paying little attention to tabular data. 
AdapTable~\cite{kim2024adaptable} studies test-time adaptation for tabular data, effectively designing a graph-based module to address label shifts and providing insightful theoretical analyses.
TabLog~\cite{ren2024tablog} is the first to examine the structure of invariant rules for tabular data in the context of test-time adaptation.
However, these stduies require the training data to be available, which cannot be applied in our \setting problem setting and is not practical in real-world scenarios. 
Therefore, our paper focuses on the fully test-time adaptation problem for tabular data, an area that remains underexplored.

\noindent \paragraph{Deep Tabular Learning.} 
Deep Tabular Learning aims to model tabular data for tasks such as classification and regression through deep learning methods. 
Unlike image and language data, the heterogeneity and high dimensionality make it difficult for models to extract spatial and semantic information. 
Recently, attention-based architectures have been introduced to the tabular data domain. FT-Transformer~\cite{Gorishniy21FTTransformer} applies a feature tokenizer to heterogeneous feature columns and learns an optimal representation in embedding space. Additionally, TabTransformer~\cite{tabtransformer}, TabNet~\cite{Arik21TabNet}, and other deep tabular models~\cite{DBLP:journals/corr/abs-2002-07971,KlambauerUMH17, GorishniyRB22, GrinsztajnOV22} are proposed for better representation of tabular data.
However, these methods typically works well in an i.i.d. setting, and may suffer from performance degradation when the test data distribution shifts.

\section{Conclusion}
In this paper, we investigate the problem of fully test-time adaptation for tabular data (\setting), an important and practically valuable issue that remains underexplored. Our observations highlight three key challenges in the \setting problem: the existence of label and covariate distribution shifts, the lack of effective data augmentation, and the sensitivity of model adaptation.
To address these challenges, we propose the \algo approach, which includes three novel modules: \emph{Confident Distribution Optimizer}, \emph{Local Consistent Weighter}, and \emph{Dynamic Model Ensembler}. Our experimental results demonstrate that the \algo approach outperforms existing FTTA methods, demonstrating its effectiveness in addressing tabular tasks.

One limitation of this paper is that the design of \algo approach lacks deep theoretical understanding and we will explore in this direction in the future to provide deep insights for the following researchers.

\section*{Acknowledgements}

This research was supported by the National Natural Science Foundation of China (Grant No. 624B2068, 62176118, and 62306133), the Key Program of Jiangsu Science Foundation (BK20243012), and the Fundamental Research Funds for the Central Universities (022114380023).
We would like to thank the reviewers for their constructive suggestions.

\bibliography{ref}

\begin{thebibliography}{41}
\providecommand{\natexlab}[1]{#1}

\bibitem[{Altman and Krzywinski(2017)}]{naomi17tabular}
Altman, N.; and Krzywinski, M. 2017.
\newblock Tabular data.
\newblock \emph{Nature Methods}, 14(4): 329--330.

\bibitem[{Alvarez-Melis and Fusi(2020)}]{alvarez2020geometric}
Alvarez-Melis, D.; and Fusi, N. 2020.
\newblock Geometric dataset distances via optimal transport.
\newblock In \emph{Advances in Neural Information Processing Systems}, 21428--21439.

\bibitem[{Arik and Pfister(2021)}]{Arik21TabNet}
Arik, S.~{\"{O}}.; and Pfister, T. 2021.
\newblock TabNet: Attentive Interpretable Tabular Learning.
\newblock In \emph{Proceedings of the 35th {AAAI} Conference on Artificial Intelligence}, 6679--6687.

\bibitem[{Badirli et~al.(2020)Badirli, Liu, Xing, Bhowmik, and Keerthi}]{DBLP:journals/corr/abs-2002-07971}
Badirli, S.; Liu, X.; Xing, Z.; Bhowmik, A.; and Keerthi, S.~S. 2020.
\newblock Gradient Boosting Neural Networks: GrowNet.
\newblock \emph{CoRR}, abs/2002.07971.

\bibitem[{Bai et~al.(2022)Bai, Zhang, Zhao, Sugiyama, and Zhou}]{bai2022adapting}
Bai, Y.; Zhang, Y.-J.; Zhao, P.; Sugiyama, M.; and Zhou, Z.-H. 2022.
\newblock Adapting to Online Label Shift with Provable Guarantees.
\newblock In \emph{Advances in Neural Information Processing Systems}.

\bibitem[{Boudiaf et~al.(2022)Boudiaf, Mueller, Ben~Ayed, and Bertinetto}]{BoudiafLAME22}
Boudiaf, M.; Mueller, R.; Ben~Ayed, I.; and Bertinetto, L. 2022.
\newblock Parameter-free Online Test-time Adaptation.
\newblock In \emph{Proceedings of the {IEEE/CVF} Conference on Computer Vision and Pattern Recognition}, 8344--8353.

\bibitem[{Chen et~al.(2022)Chen, Wang, Darrell, and Ebrahimi}]{chen19adacontrast}
Chen, D.; Wang, D.; Darrell, T.; and Ebrahimi, S. 2022.
\newblock Contrastive Test-Time Adaptation.
\newblock In \emph{Proceedings of the {IEEE/CVF} Conference on Computer Vision and Pattern Recognition}, 295--305.

\bibitem[{Ching et~al.(2018)Ching, Himmelstein, Beaulieu-Jones, Kalinin, Do, Way, Ferrero, Agapow, Zietz, Hoffman et~al.}]{ching2018opportunities}
Ching, T.; Himmelstein, D.~S.; Beaulieu-Jones, B.~K.; Kalinin, A.~A.; Do, B.~T.; Way, G.~P.; Ferrero, E.; Agapow, P.-M.; Zietz, M.; Hoffman, M.~M.; et~al. 2018.
\newblock Opportunities and obstacles for deep learning in biology and medicine.
\newblock \emph{Journal of the Royal Society Interface}, 15(141): 20170387.

\bibitem[{Fang et~al.(2022)Fang, Tang, Cui, Zhu, Li, Zhou, and Zhu}]{FangTCZLZZ22}
Fang, J.; Tang, C.; Cui, Q.; Zhu, F.; Li, L.; Zhou, J.; and Zhu, W. 2022.
\newblock Semi-Supervised Learning with Data Augmentation for Tabular Data.
\newblock In \emph{Proceedings of the 31st {ACM} International Conference on Information {\&} Knowledge Management}, 3928--3932.

\bibitem[{Fuchs and Whelton(2020)}]{Hypertension1}
Fuchs, F.~D.; and Whelton, P.~K. 2020.
\newblock High Blood Pressure and Cardiovascular Disease.
\newblock \emph{Hypertension}, 75(2): 285--292.

\bibitem[{Gardner, Popovic, and Schmidt(2023)}]{GardnerPS23}
Gardner, J.; Popovic, Z.; and Schmidt, L. 2023.
\newblock Benchmarking Distribution Shift in Tabular Data with TableShift.
\newblock In \emph{Advances in Neural Information Processing Systems}.

\bibitem[{Gong et~al.(2022)Gong, Jeong, Kim, Kim, Shin, and Lee}]{gong2022note}
Gong, T.; Jeong, J.; Kim, T.; Kim, Y.; Shin, J.; and Lee, S.-J. 2022.
\newblock {NOTE}: Robust Continual Test-time Adaptation Against Temporal Correlation.
\newblock In \emph{Advances in Neural Information Processing Systems}, 27253--27266.

\bibitem[{Gorishniy, Rubachev, and Babenko(2022)}]{GorishniyRB22}
Gorishniy, Y.; Rubachev, I.; and Babenko, A. 2022.
\newblock On Embeddings for Numerical Features in Tabular Deep Learning.
\newblock In \emph{Advances in Neural Information Processing Systems}.

\bibitem[{Gorishniy et~al.(2024)Gorishniy, Rubachev, Kartashev, Shlenskii, Kotelnikov, and Babenko}]{GorishniyRKSKB24}
Gorishniy, Y.; Rubachev, I.; Kartashev, N.; Shlenskii, D.; Kotelnikov, A.; and Babenko, A. 2024.
\newblock TabR: Tabular Deep Learning Meets Nearest Neighbors.
\newblock In \emph{Proceedings of th 12th International Conference on Learning Representations}.

\bibitem[{Gorishniy et~al.(2021)Gorishniy, Rubachev, Khrulkov, and Babenko}]{Gorishniy21FTTransformer}
Gorishniy, Y.; Rubachev, I.; Khrulkov, V.; and Babenko, A. 2021.
\newblock Revisiting Deep Learning Models for Tabular Data.
\newblock In \emph{Advances in Neural Information Processing Systems}, 18932--18943.

\bibitem[{Goyal et~al.(2022)Goyal, Sun, Raghunathan, and Kolter}]{goyal2022test}
Goyal, S.; Sun, M.; Raghunathan, A.; and Kolter, J.~Z. 2022.
\newblock Test Time Adaptation via Conjugate Pseudo-labels.
\newblock In \emph{Advances in Neural Information Processing Systems}, 6204--6218.

\bibitem[{Grinsztajn, Oyallon, and Varoquaux(2022)}]{GrinsztajnOV22}
Grinsztajn, L.; Oyallon, E.; and Varoquaux, G. 2022.
\newblock Why do tree-based models still outperform deep learning on typical tabular data?
\newblock In \emph{Advances in Neural Information Processing Systems}.

\bibitem[{Guo, Hu, and Yang(2023)}]{GuoHY23}
Guo, Y.; Hu, C.; and Yang, Y. 2023.
\newblock Predict the Future from the Past? On the Temporal Data Distribution Shift in Financial Sentiment Classifications.
\newblock In \emph{Proceedings of the 2023 Conference on Empirical Methods in Natural Language Processing}, 1029--1038.

\bibitem[{He, Girshick, and Doll{\'{a}}r(2019)}]{HeGD19}
He, K.; Girshick, R.~B.; and Doll{\'{a}}r, P. 2019.
\newblock Rethinking ImageNet Pre-Training.
\newblock In \emph{Proceedings of the 2019 {IEEE/CVF} International Conference on Computer Vision}, 4917--4926.

\bibitem[{Hein et~al.(2017)Hein, Depeweg, Tokic, Udluft, Hentschel, Runkler, and Sterzing}]{HeinDTUHRS17}
Hein, D.; Depeweg, S.; Tokic, M.; Udluft, S.; Hentschel, A.; Runkler, T.~A.; and Sterzing, V. 2017.
\newblock A benchmark environment motivated by industrial control problems.
\newblock In \emph{Proceedings of the 2017 {IEEE} Symposium Series on Computational Intelligence}, 1--8.

\bibitem[{Helton and Johnson(2011)}]{HeltonJ11}
Helton, J.~C.; and Johnson, J.~D. 2011.
\newblock Quantification of margins and uncertainties: Alternative representations of epistemic uncertainty.
\newblock \emph{Reliability Engineering \& System Safety}, 96(9): 1034--1052.

\bibitem[{Huang et~al.(2020)Huang, Khetan, Cvitkovic, and Karnin}]{tabtransformer}
Huang, X.; Khetan, A.; Cvitkovic, M.; and Karnin, Z.~S. 2020.
\newblock TabTransformer: Tabular Data Modeling Using Contextual Embeddings.
\newblock \emph{CoRR}, abs/2012.06678.

\bibitem[{Kim et~al.(2024)Kim, Kim, Woo, Yang, and Yang}]{kim2024adaptable}
Kim, C.; Kim, T.; Woo, S.; Yang, J.~Y.; and Yang, E. 2024.
\newblock AdapTable: Test-Time Adaptation for Tabular Data via Shift-Aware Uncertainty Calibrator and Label Distribution Handler.

\bibitem[{Klambauer et~al.(2017)Klambauer, Unterthiner, Mayr, and Hochreiter}]{KlambauerUMH17}
Klambauer, G.; Unterthiner, T.; Mayr, A.; and Hochreiter, S. 2017.
\newblock Self-Normalizing Neural Networks.
\newblock In \emph{Advances in Neural Information Processing Systems}, 971--980.

\bibitem[{Kolesnikov(2023)}]{Kolesnikov23wildtab}
Kolesnikov, S. 2023.
\newblock Wild-Tab: {A} Benchmark for Out-Of-Distribution Generalization in Tabular Regression.
\newblock \emph{CoRR}, abs/2312.01792.

\bibitem[{Kritzman, Page, and Turkington(2012)}]{kritzman2012regime}
Kritzman, M.; Page, S.; and Turkington, D. 2012.
\newblock Regime shifts: Implications for dynamic strategies.
\newblock \emph{Financial Analysts Journal}, 68(3): 22--39.

\bibitem[{Laine and Aila(2017)}]{LaineA17}
Laine, S.; and Aila, T. 2017.
\newblock Temporal Ensembling for Semi-Supervised Learning.
\newblock In \emph{Proceeding of the 5th International Conference on Learning Representations}.

\bibitem[{Liu et~al.(2021)Liu, Kothari, van Delft, Bellot{-}Gurlet, Mordan, and Alahi}]{LiuKDBMA21}
Liu, Y.; Kothari, P.; van Delft, B.; Bellot{-}Gurlet, B.; Mordan, T.; and Alahi, A. 2021.
\newblock {TTT++:} When Does Self-Supervised Test-Time Training Fail or Thrive?
\newblock In \emph{Advances in Neural Information Processing Systems}, 21808--21820.

\bibitem[{Niu et~al.(2022)Niu, Wu, Zhang, Chen, Zheng, Zhao, and Tan}]{NiuW0CZZT22}
Niu, S.; Wu, J.; Zhang, Y.; Chen, Y.; Zheng, S.; Zhao, P.; and Tan, M. 2022.
\newblock Efficient Test-Time Model Adaptation without Forgetting.
\newblock In \emph{Proceedings of the 39th International Conference on Machine Learning}, 16888--16905.

\bibitem[{Niu et~al.(2023)Niu, Wu, Zhang, Wen, Chen, Zhao, and Tan}]{niu2023towards}
Niu, S.; Wu, J.; Zhang, Y.; Wen, Z.; Chen, Y.; Zhao, P.; and Tan, M. 2023.
\newblock Towards Stable Test-time Adaptation in Dynamic Wild World.
\newblock In \emph{Proceedings of the 11th International Conference on Learning Representations}.

\bibitem[{Ozbayoglu, Gudelek, and Sezer(2020)}]{ozbayoglu2020deep}
Ozbayoglu, A.~M.; Gudelek, M.~U.; and Sezer, O.~B. 2020.
\newblock Deep learning for financial applications: A survey.
\newblock \emph{Applied Soft Computing}, 93: 106384.

\bibitem[{Ren et~al.(2024)Ren, Li, Chen, Rakesh, Wang, Das, and Honavar}]{ren2024tablog}
Ren, W.; Li, X.; Chen, H.; Rakesh, V.; Wang, Z.; Das, M.; and Honavar, V.~G. 2024.
\newblock TabLog: Test-Time Adaptation for Tabular Data Using Logic Rules.
\newblock In \emph{Proceedings of the 41st International Conference on Machine Learning}.

\bibitem[{Salehpour and Samadzamini(2024)}]{Salehpour2024economics}
Salehpour, A.; and Samadzamini, K. 2024.
\newblock A bibliometric analysis on the application of deep learning in economics, econometrics, and finance.
\newblock \emph{International Journal of Computer Sciences and Engineering}, 27(2): 167--181.

\bibitem[{Strack et~al.(2014)Strack, DeShazo, Gennings, Olmo, Ventura, Cios, and Clore}]{Diabetes2}
Strack, B.; DeShazo, J.~P.; Gennings, C.; Olmo, J.~L.; Ventura, S.; Cios, K.~J.; and Clore, J.~N. 2014.
\newblock Impact of HbA1c Measurement on Hospital Readmission Rates: Analysis of 70,000 Clinical Database Patient Records.
\newblock \emph{BioMed Research International}, 2014(1): 781670.

\bibitem[{Sun et~al.(2020)Sun, Wang, Liu, Miller, Efros, and Hardt}]{SunWLMEH20}
Sun, Y.; Wang, X.; Liu, Z.; Miller, J.; Efros, A.~A.; and Hardt, M. 2020.
\newblock Test-Time Training with Self-Supervision for Generalization under Distribution Shifts.
\newblock In \emph{Proceedings of the 37th International Conference on Machine Learning}, 9229--9248.

\bibitem[{Torralba and Oliva(2003)}]{torralba2003statistics}
Torralba, A.; and Oliva, A. 2003.
\newblock Statistics of natural image categories.
\newblock \emph{Network: Computation in Neural Systems}, 14(3): 391.

\bibitem[{Umpierrez et~al.(2002)Umpierrez, Isaacs, Bazargan, You, Thaler, and Kitabchi}]{Diabetes1}
Umpierrez, G.~E.; Isaacs, S.~D.; Bazargan, N.; You, X.; Thaler, L.~M.; and Kitabchi, A.~E. 2002.
\newblock Hyperglycemia: An Independent Marker of In-Hospital Mortality in Patients with Undiagnosed Diabetes.
\newblock \emph{The Journal of Clinical Endocrinology \& Metabolism}, 87(3): 978--982.

\bibitem[{Wang et~al.(2021)Wang, Shelhamer, Liu, Olshausen, and Darrell}]{WangSLOD21}
Wang, D.; Shelhamer, E.; Liu, S.; Olshausen, B.~A.; and Darrell, T. 2021.
\newblock Tent: Fully Test-Time Adaptation by Entropy Minimization.
\newblock In \emph{Proceedings of the 9th International Conference on Learning Representations}.

\bibitem[{Wang et~al.(2022)Wang, Fink, Gool, and Dai}]{WangCoTTA22}
Wang, Q.; Fink, O.; Gool, L.~V.; and Dai, D. 2022.
\newblock Continual Test-Time Domain Adaptation.
\newblock In \emph{Proceedings of the {IEEE/CVF} Conference on Computer Vision and Pattern Recognition}, 7191--7201.

\bibitem[{Wu et~al.(2021)Wu, Guo, Su, and Weinberger}]{WuGSW21}
Wu, R.; Guo, C.; Su, Y.; and Weinberger, K.~Q. 2021.
\newblock Online Adaptation to Label Distribution Shift.
\newblock In \emph{Advances in Neural Information Processing Systems}, 11340--11351.

\bibitem[{Zhou et~al.(2023)Zhou, Guo, Jia, Zhang, and Li}]{zhouz23ods}
Zhou, Z.; Guo, L.; Jia, L.; Zhang, D.; and Li, Y. 2023.
\newblock {ODS:} Test-Time Adaptation in the Presence of Open-World Data Shift.
\newblock In \emph{Proceedings of the 40th International Conference on Machine Learning}, 42574--42588.

\end{thebibliography}

\onecolumn % 切换到单栏模式

\appendix

\begin{center}
    \LARGE \bf
    Fully Test-time Adaptation for Tabular Data \\
    --- Appendix ---
\end{center}

\noindent The structure of Appendix is as follows:
\begin{itemize}
    \item Section~\ref{sec:exp-details} provides detailed experimental setup and implementations.
    \item Section~\ref{sec:exp-results} presents detailed experimental results.
    \item Section~\ref{sec:impact} discusses the border impact.
\end{itemize}

\section{Detailed Experimental Setup}
\label{sec:exp-details}

We have briefly introduced experimental setup in our main manuscript. Here, we provide detailed experimental setup and implementations in this section. 

\subsection{Details of Comparison Methods} 

To compare our \algo apporach with various FTTA algorithms, including typical FTTA methods, continual FTTA methods, and recently proposed robust FTTA methods. 
For typical TTA methods, we take two novel method into comparison: 
\begin{enumerate}
    \item[(1)] TENT~\cite{WangSLOD21} updates the model parameters with entropy minimization loss; 
    \item[(2)] EATA~\cite{NiuW0CZZT22} performs activate sample selection for adaptation and Fisher regularization for anti-forgetting to achieve strong predicting performance. 
\end{enumerate}
For continual TTA methods, we compare with:
\begin{enumerate}
    \item[(3)]  CoTTA~\cite{WangCoTTA22} eliminates error accumulated in the data stream via weight-and-augmentation averaged pseudo-labels and parameters stochastic restoration. We adopt the perturb augmentation~\cite{FangTCZLZZ22} as data augmentation methods for tabular data. 
\end{enumerate}
For robust TTA methods, we take various proposed methods into comparison: 
\begin{enumerate}
    \item[(4)]  SAR~\cite{niu2023towards} conducts sample filtering based on test entropy and update model parameters to a flat minimum to achieve well and robust performance; 
    \item[(5)]  LAME~\cite{BoudiafLAME22} modifies model output by adopting a conservative adaptation approach; 
    \item[(6)]  ODS~\cite{zhouz23ods} decouples the mixed distribution shift and then addresses covariate and label distribution shifts accordingly.
\end{enumerate}

\subsection{Implementations Details} 

In this subsection, we provide the details of backbone model, configuration of training and testing phase to enhance the reporducibility. All experiments are conducted on a Linux server with one NVIDIA GeForce RTX 3050Ti GPU. 

\paragraph{Backbone Models.}
For all experiments, we use three representative deep tabular models: MLP, Tabtransformer~\cite{tabtransformer} and FT-Transformer~\cite{Gorishniy21FTTransformer} as the backbone model. 

\paragraph{Training Phase.} 
For training the source model, we follow the TableShift benchmark~\cite{GardnerPS23} for all setting of training hyperparameters. 
Specifically, we train each backbone model with a batch size of 512 for several epochs, depending on the model’s convergence as evaluated on the validation set. The AdamW optimizer is used with a learning rate of 0.01 and a weight decay of 0.01.

\paragraph{Testing Phase.}
For test-time adaptation, we set the batch size to 512 due to the large number of samples in each test set. For the online model ensemble, we configure three base learners with different learning rates: $1e-5$, $5e-4$, and $1e-4$. 
We set the hyperparameters of \algo to $\alpha = 0.1$, $\epsilon = 0.611$, $\beta=0.3$ for all experiments to demonstrate its robustness to hyperparameters. 
The value of $\epsilon$ is computed by the function $\mathcal{E}(p) = \text{Entropy}([p, 1-p])$, with $p$ set to 0.7.
For comparison methods, we use their original hyperparameters in their paper. 
We report mean $\pm$ stdev accuracy, balanced accuracy and F1 score over three runs using different random seeds. 

\paragraph{Code and Datasets. }
The code and datasets used in this paper are available at \url{https://wnjxyk.github.io/FTTA}.

\begin{table*}[t]
    \caption{Performance of \algo approach and comparison methods on 6 datasets using TabTransformer backbone model. The best performance is in \textbf{bold}.}
    \label{tab:exps-tab}
    \centering
    \resizebox{\linewidth}{!}{
    \begin{tabular}{l|ccc|ccc|ccc}
    \bottomrule
    \toprule
    \multirow{2}{*}{Method} & \multicolumn{3}{c|}{HELOC} & \multicolumn{3}{c|}{ANES} &\multicolumn{3}{c}{Health Ins.}  \\ \cmidrule{2-10} 
      & Acc. & Balanced Acc. & F1  & Acc. & Balanced Acc. & F1  & Acc. & Balanced Acc. & F1 \\
    \midrule
    Non-Adaptation	 & $55.66\pm1.34$	 & $59.60\pm1.11$	 & $44.30\pm3.08$	 & $78.95\pm0.27$	 & $75.30\pm0.45$	 & \bm{$84.23\pm0.12$}	& $65.35\pm1.08$ & $70.11\pm0.67$ & $65.88\pm1.63$\\
    TENT	 & $56.37\pm1.55$	 & $59.60\pm1.11$	 & $50.34\pm5.91$	 & $78.82\pm0.38$	 & $75.18\pm0.56$	 & $84.14\pm0.19$	& $65.22\pm1.18$ & $70.07\pm0.70$ & $65.62\pm1.82$\\
    EATA	 & $56.37\pm1.55$	 & $57.04\pm4.04$	 & $52.57\pm4.42$	 & $78.82\pm0.38$	 & $75.18\pm0.56$	 & $84.14\pm0.19$	& $65.35\pm1.08$ & \bm{$70.11\pm0.67$} & $65.88\pm1.63$\\
    LAME	 & $51.71\pm4.68$	 & $59.56\pm1.51$	 & $44.30\pm6.64$	 & $78.57\pm0.49$	 & $74.77\pm0.80$	 & $84.04\pm0.20$	& $65.35\pm1.08$ & \bm{$70.11\pm0.67$} & $65.88\pm1.63$\\
    CoTTA	 & $56.37\pm1.55$	 & $59.60\pm1.11$	 & $50.34\pm5.91$	 & $78.82\pm0.38$	 & $75.18\pm0.56$	 & $84.14\pm0.19$	& $65.35\pm1.08$ & $70.11\pm0.67$ & $65.88\pm1.63$\\
    ODS	 & $52.19\pm4.15$	 & $56.77\pm3.56$	 & $44.70\pm6.56$	 & $78.48\pm0.70$	 & $74.69\pm1.09$	 & $83.97\pm0.31$	& $57.14\pm5.22$ & $64.75\pm3.23$ & $51.33\pm10.03$\\
    SAR	 & $56.37\pm1.55$	 & $59.60\pm1.11$	 & $50.34\pm5.91$	 & $78.82\pm0.38$	 & $75.18\pm0.56$	 & $84.14\pm0.19$	& $65.35\pm1.08$ & $70.11\pm0.67$ & $65.88\pm1.63$\\
    \algo	 & \bm{$65.01\pm1.09$}	 & \bm{$64.34\pm1.05$}	 & \bm{$69.06\pm1.11$}	 & \bm{$80.00\pm0.19$}	 & \bm{$79.07\pm0.14$}	 & $83.32\pm0.25$	& \bm{$69.95\pm0.44$} & $61.27\pm0.93$ & \bm{$79.77\pm0.10$}\\
    \midrule \midrule
    \multirow{2}{*}{Method} & \multicolumn{3}{c|}{ASSIST} & \multicolumn{3}{c|}{DIABETE} &\multicolumn{3}{c}{Hypertension} \\ \cmidrule{2-10} 
      & Acc. & Balanced Acc. & F1  & Acc. & Balanced Acc. & F1  & Acc. & Balanced Acc. & F1 \\
    \midrule
    Non-Adaptation	 & $49.04\pm4.23$	 & $53.07\pm2.12$	 & $58.92\pm2.34$	 & $60.85\pm0.32$	 & $60.59\pm0.34$	 & $50.00\pm1.44$	& $54.87\pm0.89$ & $59.24\pm0.63$ & $46.25\pm2.16$\\
    TENT	 & $47.83\pm3.39$	 & $53.11\pm2.18$	 & $61.27\pm0.32$	 & $60.85\pm0.06$	 & $60.59\pm0.06$	 & $50.00\pm0.11$	& $41.74\pm0.01$ & $50.12\pm0.01$ & $0.78\pm0.06$\\
    EATA	 & $45.28\pm0.16$	 & $51.44\pm0.14$	 & $61.40\pm0.41$	 & $60.85\pm0.06$	 & $60.59\pm0.05$	 & $50.00\pm0.09$	& $54.86\pm0.89$ & $59.24\pm0.63$ & $46.24\pm2.15$\\
    LAME	 & $45.12\pm3.46$	 & $51.30\pm2.20$	 & $61.40\pm0.41$	 & $61.19\pm0.04$	 & $60.96\pm0.05$	 & $51.76\pm0.31$	& $54.87\pm0.89$ & $59.24\pm0.63$ & $46.25\pm2.16$\\
    CoTTA	 & $45.51\pm0.39$	 & $51.64\pm0.34$	 & \bm{$61.56\pm0.16$}	 & $60.85\pm0.06$	 & $60.59\pm0.06$	 & $50.00\pm0.11$	& $54.87\pm0.89$ & $59.24\pm0.63$ & $46.25\pm2.16$\\
    ODS	 & $45.12\pm3.46$	 & $51.30\pm2.20$	 & $61.40\pm0.41$	 & $61.23\pm0.11$	 & $60.99\pm0.12$	 & $51.83\pm0.43$	& $54.87\pm0.89$ & $59.24\pm0.63$ & $46.25\pm2.16$\\
    SAR	 & $45.12\pm3.46$	 & $51.30\pm2.20$	 & $61.40\pm0.41$	 & $60.85\pm0.07$	 & $60.59\pm0.06$	 & $50.00\pm0.11$	& $54.87\pm0.89$ & $59.24\pm0.63$ & $46.25\pm2.16$\\
    \algo	 & \bm{$54.81\pm1.80$}	 & \bm{$55.33\pm0.59$}	 & $53.20\pm2.62$	 & \bm{$61.74\pm0.27$}	 & \bm{$61.64\pm0.27$}	 & \bm{$57.95\pm0.38$}	& \bm{$62.88\pm0.25$} & \bm{$60.72\pm0.27$} & \bm{$69.82\pm0.67$}\\
    \bottomrule
    \toprule
    \end{tabular}}
\end{table*}

\begin{table*}[t]
    \caption{Performance of \algo approach and comparison methods on 6 datasets using FT-Transformer backbone model. The best performance is in \textbf{bold}.}
    \label{tab:exps-trans}
    \centering
    \resizebox{\textwidth}{!}{
    \begin{tabular}{l|ccc|ccc|ccc}
    \bottomrule
    \toprule
    \multirow{2}{*}{Method} & \multicolumn{3}{c|}{HELOC} & \multicolumn{3}{c|}{ANES} &\multicolumn{3}{c}{Health Ins.}  \\ \cmidrule{2-10} 
      & Acc. & Balanced Acc. & F1  & Acc. & Balanced Acc. & F1  & Acc. & Balanced Acc. & F1  \\
    \midrule
    Non-Adaptation	 & $46.26\pm1.05$	 & $52.48\pm0.78$	 & $13.32\pm4.50$	 & \bm{$75.47\pm1.31$}	 & $71.50\pm2.06$	 & \bm{$81.80\pm0.45$}	& $58.33\pm4.05$ & \bm{$65.44\pm2.56$} & $54.06\pm7.72$\\
    TENT	 & $44.98\pm1.83$	 & $51.45\pm1.39$	 & $8.11\pm7.80$	 & $63.02\pm0.89$	 & $54.52\pm1.15$	 & $76.19\pm0.40$	& $36.44\pm0.03$ & $50.05\pm0.02$ & $0.24\pm0.11$\\
    EATA	 & $45.95\pm1.00$	 & $52.23\pm0.74$	 & $12.27\pm4.38$	 & $74.65\pm1.66$	 & $70.16\pm2.49$	 & $81.51\pm0.67$	& $57.40\pm4.35$ & $64.86\pm2.79$ & $52.23\pm8.49$\\
    LAME	 & $43.14\pm0.04$	 & $50.03\pm0.03$	 & $0.20\pm0.14$	 & $75.37\pm1.34$	 & $71.35\pm2.12$	 & $81.73\pm0.48$	& $59.08\pm4.57$ & $65.44\pm2.60$ & $55.91\pm9.04$\\
    CoTTA	 & $46.26\pm1.05$	 & $52.48\pm0.78$	 & $10.67\pm7.65$	 & \bm{$75.47\pm1.31$}	 & $71.50\pm2.06$	 & \bm{$81.80\pm0.45$}	& $58.33\pm4.05$ & \bm{$65.44\pm2.56$} & $54.06\pm7.72$\\
    ODS	 & $43.14\pm0.04$	 & $50.03\pm0.03$	 & $0.20\pm0.14$	 & $75.41\pm1.39$	 & $71.41\pm2.15$	 & $81.75\pm0.54$	& $59.99\pm3.45$ & $65.54\pm1.53$ & $58.37\pm6.92$\\
    SAR	 & $43.30\pm0.00$	 & $50.20\pm0.00$	 & $30.60\pm0.00$	 & \bm{$75.47\pm1.31$}	 & $71.50\pm2.06$	 & \bm{$81.80\pm0.45$}	& $58.33\pm4.05$ & \bm{$65.44\pm2.56$} & $54.06\pm7.72$\\
    \algo	 & \bm{$65.09\pm0.27$}	 & \bm{$63.76\pm0.39$}	 & \bm{$70.49\pm1.25$}	 & $72.58\pm0.18$	 & \bm{$74.40\pm0.28$}	 & \bm{$73.73\pm0.21$}	& $63.78\pm0.07$ & $50.26\pm0.11$ & \bm{$77.82\pm0.03$}\\
    \midrule \midrule
    \multirow{2}{*}{Method} & \multicolumn{3}{c|}{ASSIST} & \multicolumn{3}{c|}{DIABETE} &\multicolumn{3}{c}{Hypertension} \\ \cmidrule{2-10} 
      & Acc. & Balanced Acc. & F1  & Acc. & Balanced Acc. & F1  & Acc. & Balanced Acc. & F1  \\
    \midrule
    Non-Adaptation	 & $58.32\pm0.07$	 & $62.99\pm0.09$	 & \bm{$67.63\pm0.09$}	 & $60.86\pm0.17$	 & $60.67\pm0.19$	 & $53.21\pm1.34$	& $58.88\pm0.28$ & $61.84\pm0.14$ & $55.71\pm0.79$\\
    TENT	 & $58.25\pm0.07$	 & $62.91\pm0.09$	 & $67.57\pm0.09$	 & $59.97\pm0.33$	 & $59.68\pm0.35$	 & $47.12\pm1.48$	& $47.01\pm6.39$ & $53.88\pm4.52$ & $18.83\pm20.89$\\
    EATA	 & $48.04\pm0.25$	 & $53.85\pm0.19$	 & $62.60\pm0.07$	 & $51.33\pm0.47$	 & $50.74\pm0.46$	 & $3.47\pm2.12$	& $58.84\pm0.27$ & $61.82\pm0.13$ & $55.62\pm0.78$\\
    LAME	 & $56.54\pm2.48$	 & $62.98\pm0.06$	 & \bm{$67.63\pm0.07$}	 & $60.52\pm0.19$	 & $60.28\pm0.20$	 & $50.23\pm0.91$	& $58.78\pm0.27$ & $61.78\pm0.15$ & $55.47\pm0.74$\\
    CoTTA	 & $58.25\pm0.07$	 & $62.91\pm0.09$	 & $67.57\pm0.09$	 & $60.64\pm0.14$	 & $60.40\pm0.15$	 & $50.67\pm0.71$	& $58.88\pm0.28$ & $61.84\pm0.14$ & $55.71\pm0.79$\\
    ODS	 & $57.39\pm1.19$	 & $62.16\pm1.04$	 & $67.14\pm0.58$	 & $59.60\pm0.52$	 & $59.29\pm0.55$	 & $45.53\pm2.23$	& $58.77\pm0.26$ & $61.78\pm0.14$ & $55.45\pm0.72$\\
    SAR	 & $58.25\pm0.07$	 & $62.91\pm0.09$	 & $67.57\pm0.09$	 & $60.63\pm0.14$	 & $60.39\pm0.15$	 & $50.67\pm0.71$	& $59.64\pm0.65$ & $62.24\pm0.29$ & $57.52\pm1.63$\\
    \algo	 & \bm{$58.62\pm0.24$}	 & \bm{$63.13\pm0.22$}	 & $67.55\pm0.14$	 & \bm{$61.16\pm0.09$}	 & \bm{$61.17\pm0.09$}	 & \bm{$61.25\pm0.31$}	& \bm{$62.81\pm1.22$} & \bm{$62.50\pm0.57$} & \bm{$66.52\pm3.98$}\\
    \bottomrule
    \toprule
    \end{tabular}}
\end{table*}

\subsection{Dataset Details}

We conduct experiments on tabular datasets with natural distribution shifts from the TableShift benchmark~\cite{GardnerPS23}. 
We selected six datasets suffering notable performance degradation under distribution shifts and containing sufficient sample sizes for adaptation, to demonstrate the effectiveness of our \algo approach.
Overall, we conduct experiments with samples from 10K to 5M and features from 26 to 365, including HELOC, ANES, ASSISTMENTS, DIABETES, Hypertension and Health Ins, which are widely used in various applications. 
We give the details of each dataset as follows. 

\paragraph{HELOC.} 
The Home Equity Line of Credit (HELOC) is a part of credit secured by the applicant's home. A HELOC provides access to a revolving credit line with a lower interest rate than other types of loans. 
To assess an applicant's suitability for a HELOC, a lender evaluates an applicants' financial background to predict whether a given applicant islikely to repay a line of credit and, if so, how much credit should be extended.
In order to accurate credit risk predictions for their overall utility for both lenders and borrowers and achieve equal treatment, HELOC dataset which contains 10,459 samples need to predict whether a consumer is 90 days past due or worse at least once over a period of 24 months using 38 features varying from financial activity to credit inquiries.

\paragraph{ANES.} Understanding participation in elections is a critical task for policymakers, politicians, and those with an interest in democracy.  
Predicting which individuals will vote in an electio, is widely acknowledged as critical to polling and campaigning in U.S. politics. 
For better understanding individual activity in presidential election, ANES dataset predict whether an individual will vote in the U.S presidential election using 365 features including voting behavior, elections, public opinion and attitudes.
The ANES dataset includes 60,377 samples.

\paragraph{ASSIST.} ASSISTMENTS dataset in education field (abbreviated as ASSIST). The ASSISTments tutoring platform is a free, web-based, data-driven tutoring platform for students in grades 3-12. 
ASSISTMENTS dataset contains affect predictions such as such as boredom, confusion, frustration, and engaged problem-solving behavior on students who use the ASSISTMENTS tutoring platform. 
The numbers of features and samples in ASSISTMENTS dataset are 26 and 2,667,776. 

\paragraph{DIABETES.} Effective management and treatment of diabetic patients admitted to the hospital can have a significant impact on their health outcomes~\cite{Diabetes1}. 
Several factors can affect the quality of treatment patients receive~\cite{Diabetes2}. One of the costliest and potentially most adverse outcomes after a patient is released from the hospital is for that patient to be readmitted soon after their initial release. Thus, predicting the readmission of patients is a priority from both a medical and economic perspective.
DIABETE dataset which contains 1,444,176 samples predict whether a diabetic patient is readmitted to the hospital within 30 days of their initial release using 183 related features.

\paragraph{Hypertension.} Hypertension, or systolic blood pressure (typically systolic pressure 130 mm Hg or higher or diastolic 80 or higher) affects nearly half of Americans.
When left untreated, hypertension is associated with the strongest evidence for causation of all risk factors for heart attack and other cardiovascular disease~\cite{Hypertension1}. As a result, it is important to predict blood pressure accurately and efficitively.
Hypertension dataset has a goal to achieve efficitive blood pressure measurement and increase the prediction accuracy. Hypertension dataset contains 846,781 samples with 100 features related to several risk factors for hypertension.

\paragraph{Health Ins.} Public health insurance makes a significant performance in providing affordable and accessible medical care for individuals. 
A high level of health insurance ownership is important for the healthy development of the individual. So, it is important to raise the rate of owning health insurance.
Health Ins. dataset is related to public coverage field and the goal is to predict whether an individual is covered by public health insurance using 135 features. The number of samples in Health Ins. dataset is 5,916,565.

\section{Detailed Experimental Results}

\label{sec:exp-results}

We neglect the detailed experimental results of TabTransformer~\cite{tabtransformer} and FT-Transformer~\cite{Gorishniy21FTTransformer} backbone models in our main manuscript due to the space limit. 
The neglected results are presented in Tab.~\ref{tab:exps-trans} and Tab.~\ref{tab:exps-tab}. 
The results show that our \algo approach gives the best performance on majority of datasets, demonstrating the effectiveness of our approach. In the remaining cases, \algo approach achieves competitive performance compared to the best-performing method.

\section{Broader Impact}

\label{sec:impact}

This paper aims to advance the field of deep tabular learning by addressing the negative effects of distribution shifts that occur during the testing phase. We believe our work has many potential societal impacts, the majority of which are positive and none of which need to be highlighted here.

\end{document}